\pdfoutput=1

\documentclass[11pt]{article}

\usepackage[final]{EMNLP2023}

\usepackage{subcaption}

\usepackage{booktabs} 
\usepackage{multirow}

\usepackage{siunitx} 
\usepackage{graphicx}
\usepackage[utf8]{inputenc}
\usepackage{hyperref}
\usepackage{babel,blindtext}
\usepackage[T1]{fontenc} 
\usepackage{amsmath,amssymb,mathptmx,amsfonts}
\usepackage{algorithmic}
\usepackage{textcomp}

\usepackage[authormarkup=none, final]{changes}

\usepackage{makecell}

\usepackage[export]{adjustbox}

\usepackage{multirow}
\usepackage{multicol}
\usepackage{verbatim}
\usepackage[inline]{enumitem}
\usepackage{dirtytalk}
\usepackage{pgfplots}
\usepackage{tikz}
\usetikzlibrary{shapes,backgrounds}
\usetikzlibrary{shapes,arrows,calc}
\usepgfplotslibrary{statistics}
\usepackage{url}
\usepackage{float}
\usepackage{comment}
\usetikzlibrary{tikzmark}
\usetikzlibrary{calc}
\usepackage{caption}
\usepackage{cleveref}
\usepackage{subcaption}
\usepackage{alphalph}
\usepackage{balance}

\usepackage[scaled]{helvet} 
\usepackage{courier}  

\usepackage{times}
\usepackage{latexsym}
\usepackage{subcaption,graphicx}
\usepackage{microtype}
\usepackage{tcolorbox}
\usepackage{url}
\usepackage{mathtools}
\usepackage{cleveref}
\usepackage[perpage]{footmisc}

\usepackage{makecell}
\usepackage{lscape} 
\usepackage{microtype}
\usepackage{enumitem,kantlipsum}
\usepackage{adjustbox}
\usepackage{xcolor}
\usepackage{pifont}
\restylefloat{table}
\crefformat{section}{\S#2#1#3} 
\crefformat{subsection}{\S#2#1#3}
\title{Thesis Distillation: \\
Investigating The Impact of Bias in NLP Models on Hate Speech Detection}

\author{Fatma Elsafoury \\
 Fraunhofer Research Institute (FOKUS), Berlin, Germany \\
 \texttt{fatma.elsafoury@fokus.fraunhofer.de} \\}

\begin{document}
\maketitle

\begin{abstract}
This paper is a summary of the work done in my PhD thesis. Where I investigate the impact of bias in NLP models on the task of hate speech detection from three perspectives: explainability, offensive stereotyping bias, and fairness. Then, I discuss the main takeaways from my thesis and how they can benefit the broader NLP community. Finally, I discuss important future research directions. The findings of my thesis suggest that the bias in NLP models impacts the task of hate speech detection from all three perspectives. And that unless we start incorporating social sciences in studying bias in NLP models, we will not effectively overcome the current limitations of measuring and mitigating bias in NLP models.
\end{abstract}
\section {Introduction}
Hate speech on social media has severe negative impacts, not only on its victims \cite{sticca2013longitudinal} but also on the moderators of social media platforms \cite{Roberts2019}. This is why it is crucial to develop tools for automated hate speech detection. These tools should provide a safer environment for individuals, especially for members of marginalized groups, to express themselves online. However, recent research shows that current hate speech detection models falsely flag content written by members of marginalized communities, as hateful \cite{sap2019,Dixon2018,Mchangama2021}. Similarly, recent research indicates that there are social biases in natural language processing (NLP) models \cite{Garg2017, nangia-etal-2020-crows,kurita-etal-2019-measuring, ousidhoum-etal-2021-probing, nozza-etal-2021-honest, nozza-etal-2022-measuring}. 

Yet, the impact of these biases on the task of hate speech detection has been understudied. In my thesis, I identify and study three research problems: 1) the impact of bias in NLP models on the performance and explainability of hate speech detection models; 2) the impact of the imbalanced representation of hateful content on the bias in NLP models; and 3) the impact of bias in NLP models on the fairness of hate speech detection models. 

Investigating and understanding the impact of bias in NLP on hate speech detection models will help the NLP community to develop more reliable, effective, and fair hate speech detection models. My research findings can be extended to the general task of text classification. Similarly, understanding the origins of bias in NLP models and the limitations of the current research on bias and fairness in NLP models, will help the NLP community develop more effective methods to expose and mitigate the bias in NLP models.

In my thesis and this paper, I, first, critically review the literature on hate speech detection (\cref{sec:hate_speech_survey}) and bias and fairness in NLP models (\cref{sec:bias_survey}). Then, I address the identified research problems in hate speech detection, by investigating the impact of bias in NLP models on hate speech detection models from three perspectives: 1) the explainability perspective (\cref{sec:explainabiity}), where I address the first research problem and investigate the impact of bias in NLP models on their performance of hate speech detection and whether the bias in NLP models explains their performance on hate speech detection; 2) the offensive stereotyping bias perspective (\cref{sec:sos}), where I address the second research problem and investigate the impact of imbalanced representations and co-occurrences of hateful content with marginalized identity groups on the bias of NLP models; and 3) the fairness perspective (\cref{sec:fairness}), where I address the third research problem and investigate the impact of bias in NLP models on the fairness of the task of hate speech detection. For each research problem, I summarize the work done to highlight its main findings, contributions, and limitations. Thereafter, I discuss the general takeaways from my thesis and how it can benefit the NLP community at large (\cref{sec:important_lessons}). Finally, I present directions for future research (\cref{sec:future_work}).

The findings of my thesis suggest that the bias in NLP models has an impact on hate speech detection models from all three perspectives. This means that we need to mitigate the bias in NLP models so that we can ensure the reliability of hate speech detection models. Additionally, I argue that the limitations and criticisms of the currently used methods to measure and mitigate bias in NLP models are direct results of failing to incorporate relevant literature from social sciences. I build on my findings on hate speech detection and provide a list of actionable recommendations to improve the fairness of the task of text classification as a short time solution. For a long-term solution to mitigate the bias in NLP models, I propose a list of recommendations to address bias in NLP models by addressing the underlying causes of bias from a social science perspective.

\section{Survey: Hate speech}
\label{sec:hate_speech_survey}
In \citet{elsafoury2021-survey}, I provide a comprehensive literature review on hate speech and its different forms. Furthermore, I review the literature of hate speech detection for different methods proposed in the literature accomplishing every step in the text classification pipeline. Then, I point out the limitations and challenges of the current research on hate speech detection.

\textbf{The main contributions} of this survey are: 1) There are different definitions and forms of hate speech. One of the main limitations of current studies on hate speech detection, is the lack of distinction between hate speech and other concepts like cyberbullying.  2) There are many resources of hate speech related datasets in the literature, that allow the development of new hate speech detection models. However, these datasets have many limitations, including limited languages, biased annotations, class imbalances, and user distribution imbalances. 3) One of the main limitations of the current research on hate speech detection, is the lack of understanding how it is impacted by the bias in NLP models. This limitation is what I aim to address in my thesis.

\paragraph{Limitations:}
One of the main limitations of this survey, is that it focuses on hate speech detection only as a supervised text classification task. However, recent studies propose a framework to automate and enforce moderation policies, instead of training machine learning models to detect hate speech \cite{DBLP:journals/tacl/CalabreseRL22}. Similarly, this review focuses on hate speech datasets that are collected only from social media platforms. However, recently, generative models have become more popular and started to be used in generating hate speech related datasets \cite{hartvigsen-etal-2022-toxigen}. 
 \begin{figure}
    \centering
\includegraphics[width=0.45\textwidth]{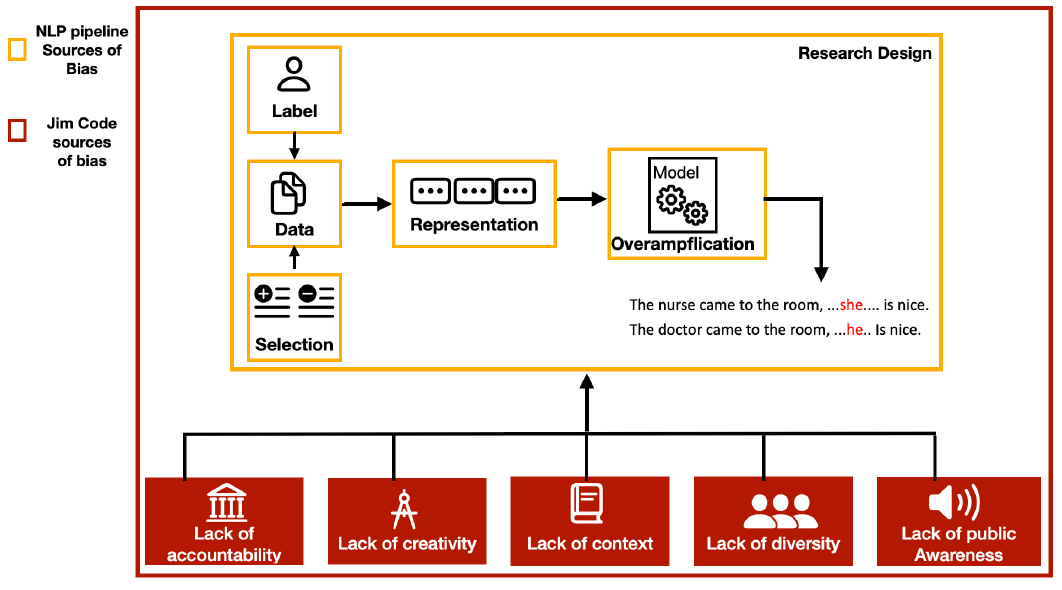}
    \caption {\small{The sources of bias in supervised NLP models. Jim code refers to social science perspective \cite{benjamin2019race}.}}
    \label{fig:sources_of_bias}
\end{figure}
\section{Survey: Bias and Fairness in NLP}
\label{sec:bias_survey}
In \citet{elsafoury2023origins}, I review the literature on the definitions of bias and fairness in NLP models. Additionally, I review the literature on the origins of bias in NLP models from two perspectives: 1) NLP pipeline as discussed in \citet{shah-etal-2020-predictive, hovy2021five},  and 2) social sciences and critical race theory as discussed in \citet{benjamin2019race, Broussard2023, nobel2018}. 

There are many definitions of the term \textit{bias}. The normative definition of bias, 
in cognitive science, is: \textit{``Behaving according to some cognitive priors and presumed realities that might not be true at all''} \citep{munoz2021}. 
And
the statistical definition of bias is ``\textit{A systematic distortion in the sampled data that compromises its representatives''} \citep{olteanu2019}. The statistical definition of bias is the one used in this thesis.

In this work, I argue that the sources of bias in the NLP pipeline originate in the social sciences and that they are direct results of the sources of bias from the social science (Jim code) perspective as shown in \Cref{fig:sources_of_bias}.

\textbf{The main contribution} of this literature review is reviewing the sources of bias in NLP models from the social science perspective as well as the NLP perspective. This survey points out the limitations of the currently used methods to measure and mitigate bias in NLP models.  It also suggests that these limitations are direct results of the lack of inclusion of social science literature in the development of methods that quantify and mitigate bias in NLP. Finally, I share a list of actionable suggestions and recommendations with the NLP community on how to mitigate the limitations discussed in studying bias in NLP (\cref{sec:important_lessons}).

\paragraph{Limitations:}
One main limitation of this survey is that it reviews the literature on the sources of bias in the NLP pipeline, only in supervised models. Unsupervised NLP models might have different sources of bias. The second limitation is regarding the reviewed literature on the sources of bias in social sciences, where I rely mainly on three books \textit{Algorithms of Oppression: How Search Engines Reinforce Racism} by Safiya Nobel \cite{nobel2018}, \textit{Race after Technology: Abolitionist Tools for the New Jim Code} by Ruha Benjamin \citet{benjamin2019race}, and \textit{More than a glitch: Confronting race, gender, and ability bias in tech} by Meredith Broussard \cite{Broussard2023}. A more comprehensive literature review to review studies that investigate the direct impact of social causes on bias in NLP would be important future work. However, to the best of my knowledge, this area is currently understudied. 

In the next sections, I address the understudied impact of bias in NLP models on hate speech detection models. I investigate that impact from the following perspectives.

\section{The explainability perspective}
\label{sec:explainabiity}
\begin{figure}[h]
    \centering
        \includegraphics[width=0.45\textwidth]{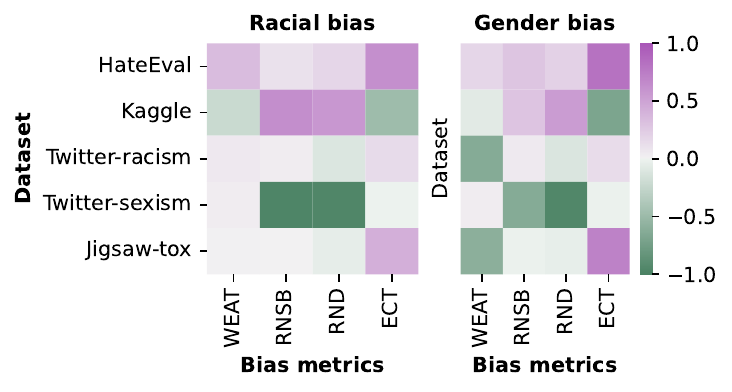}
    \caption{\small{Heatmap of the Pearson correlation coefficients between the performance (F1-scores) of SWE on the different hate speech datasets and the social bias scores.}}
\label{fig:heatmap_correlation_bewteen_socialbias_ate_speech_performance_WE}
    \end{figure}

For this perspective, I investigate the performance of different hate speech detection models and  whether the bias in NLP models explains their performance on the task of hate speech detection. To achieve that, I investigate two sources of bias: 
\begin{enumerate}[wide=0pt]
\itemsep0em 
    \item \textbf{Bias introduced by pre-training:} where I investigate the role that pre-training a language model has on the model's performance, especially when we don't know the bias in the pre-training dataset. I investigate the explainability of the performance of contextual word embeddings, also known as language models (LMs),  on the task of hate speech detection. 
    I analyze BERT's attention weights and BERT's feature importance scores. I also investigate the most important part of speech (POS) tags that BERT relies on for its performance. The results of this work suggest that pre-training BERT results in a syntactical bias that impacts its performance on the task of hate speech detection \cite{elsafoury2021}.

    Based on these findings, I investigate whether the social bias resulting from pre-training contextual word embeddings explains their performance on hate speech detection in the same way syntactical bias does. I inspect the social bias in  three LMs (BERT (base and large) \cite{devlin-etal-2019-bert}, ALBERT (base and xx-large) \cite{albert}, and ROBERTA (base and large) \cite{Roberta}) using three different bias metrics, CrowS-Pairs \cite{nangia-etal-2020-crows}, StereoSet \cite{nadeem-etal-2021-stereoset}, and SEAT \cite{DBLP:conf/naacl/MayWBBR19}, to measure gender, racial and religion biases. First, I investigate whether large models are more socially biased than base models. The Wilcoxon statistical significance test \cite{zimmerman1993relative} indicates that there is no statistical significant difference between the bias in base  and large models in BERT and RoBERTa, unlike the findings of \cite{nadeem-etal-2021-stereoset}. However, there is a significant difference between the base and xx-large ALBERT. These results suggest that large models are not necessarily more biased than base models, but if the model size gets even bigger, like ALBERT-xx-large, then the models might get significantly more biased. Since there is no significant difference between the base and large models, I only use base LMs in the rest of  the thesis.
    
    Then, I follow the work of \cite{steed-etal-2022-upstream, goldfarb-tarrant-etal-2021-intrinsic} and use correlation as a measure of the impact of bias on the performance of the task of hate speech detection. The Pearson's correlation coefficients between the bias scores of the different models and the F1-scores of the different models on the used five hate-speech-related datasets are inconsistently positive as shown in \Cref{fig:heatmap_correlation_bewteen_socialbias_ate_speech_performance}. However, due to the limitations of the metric used to measure social bias, as explained in \citet{Blodgett-etal-2021-norweigan-salmon}, the impact of the social bias in contextual word embeddings on their performance on the task of hate speech detection remains inconclusive.
 \begin{figure}
    \centering
        \includegraphics[width=0.45\textwidth]{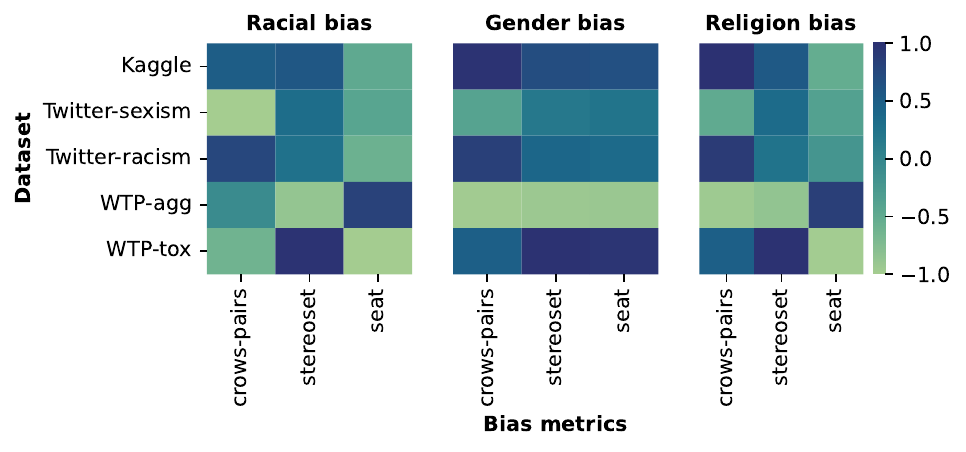}
    \caption{\small{Heatmap of the Pearson correlation coefficients between the performance (F1-scores) of LMS on the different hate speech datasets and the social bias scores.}}
    \label{fig:heatmap_correlation_bewteen_socialbias_ate_speech_performance}
    \end{figure}
    \item \textbf{Bias in pre-training datasets:} Where I investigate the impact of using NLP models pre-trained on data collected from social media platforms like Urban dictionary and 4 \& 8 Chan, which are famous for having sexist and racist posts \cite{DBLP:journals/corr/abs-1712-08647, DBLP:journals/corr/abs-2001-07487}. I investigate the performance of two groups of static word embeddings (SWE) on hate speech detection. The first group, social-media-based, pre-trained on biased datasets that contain hateful content. This group consists of Glove-Twitter \cite{10.1007/978-3-030-36687-2_77}, Urban dictionary (UD) \cite{DBLP:conf/lrec/WilsonMMGT20}, and 4\& 8 Chan (chan) \cite{DBLP:journals/corr/abs-2005-06946} word embeddings. The second group of word embeddings, informational-based, is pre-trained on informational data collected from Wikipedia and Google New platforms. This group contains the word2vec \cite{word2vec} and Glove-WK word \cite{DBLP:conf/emnlp/PenningtonSM14} embeddings. SWE in this part of the work because there are SWE that are pre-trained on datasets collected from social media platforms like urban dictionary, and 4 \&8 Chan. First, I investigate the ability of the five different word embeddings, to categorize offensive terms in the Hurtlex lexicon. Then, I investigate the performance of Bi-LSTM model with an un-trainable embeddings layer of the five word embeddings on the used five hate-speech-related datasets. The results indicate that the word embeddings that are pre-trained on biased datasets social-media-based, outperform the other word embeddings that are trained on informational data, informational-based on the tasks of offenses categorization and hate speech detection \cite{elsafoury-etal-2022-comparative}.

    Based on these findings, I inspect the impact of social bias, gender, and racial, in the SWE on their performance on the task of hate speech detection. To measure the social bias in the SWE, I use the following metrics from the literature:  WEAT \cite{Caliskan2017}, RNSB \cite{sweeney2019}, RND \cite{Garg2017}, and ECT \cite{dev2019}. Then, I use Pearson's correlation to investigate whether the social bias in the word embeddings explains their performance on the task of hate speech detection. Similar to LMs, the results indicate an inconsistent positive correlation between the bias scores and the F1-sores of the Bi-LSTM model using the different word embeddings as shown in \Cref{fig:heatmap_correlation_bewteen_socialbias_ate_speech_performance_WE}. This lack of positive correlation could be due to limitations in the used metrics to measure social bias in SWE \cite{antoniak2021}. These results suggest that the impact of the social bias in the SWE on the performance of the task of hate speech detection is inconclusive.
\end{enumerate}
\paragraph{Contributions:}
The main findings and contributions of the explainability perspective can be summarized as: \textbf{1)} The results provide evidence that the syntactical bias in contextual word embeddings, resulting from pre-training, explains their performance on the task of hate speech detection. \textbf{2)} The results suggest that pre-training static word embeddings on biased datasets from social-media-based sources improves and might explain the performance of the word embeddings on the task of hate speech detection.  \textbf{3)} For both static and contextual word embeddings, there is no strong evidence that social bias explains the performance of hate speech detection models. However, due to the limitations of the methods used to measure social bias in both static and contextual word embeddings, this finding remains inconclusive.
\paragraph{Limitations:}
one of the main limitations of this work is using social bias metrics from the literature, which have their limitations as argued in \citet{Blodgett-etal-2021-norweigan-salmon,antoniak2021}. Additionally, the work done here, is limited to hate speech datasets that are in English. Similarly, the social bias inspected in the different word embeddings is based on Western societies, where the marginalized groups might be different in different societies. It is also important to mention that the findings of this work are limited to the used datasets and models and might not generalize to other models or datasets.

\section{The offensive stereotyping bias perspective}
\label{sec:sos}
In \citet{elsafoury_sos_2022, elsafoury2023sosLM}, I investigate how the hateful content on social media and other platforms that are used to collect data and pre-train NLP models, is being encoded by those NLP models to form systematic offensive stereotyping (SOS) bias against marginalized groups of people. Especially with imbalanced representation and co-occurrence of the hateful content with the marginalized identity groups. I introduce the systematic offensive stereotyping (SOS) bias and formally define it as ``\textit{A systematic association in the word embeddings between profanity and marginalized groups of people.}'' \cite{DBLP:conf/acl/Elsafoury22}.

\begin{figure}[h]
    \centering
        \includegraphics[width=0.45\textwidth]{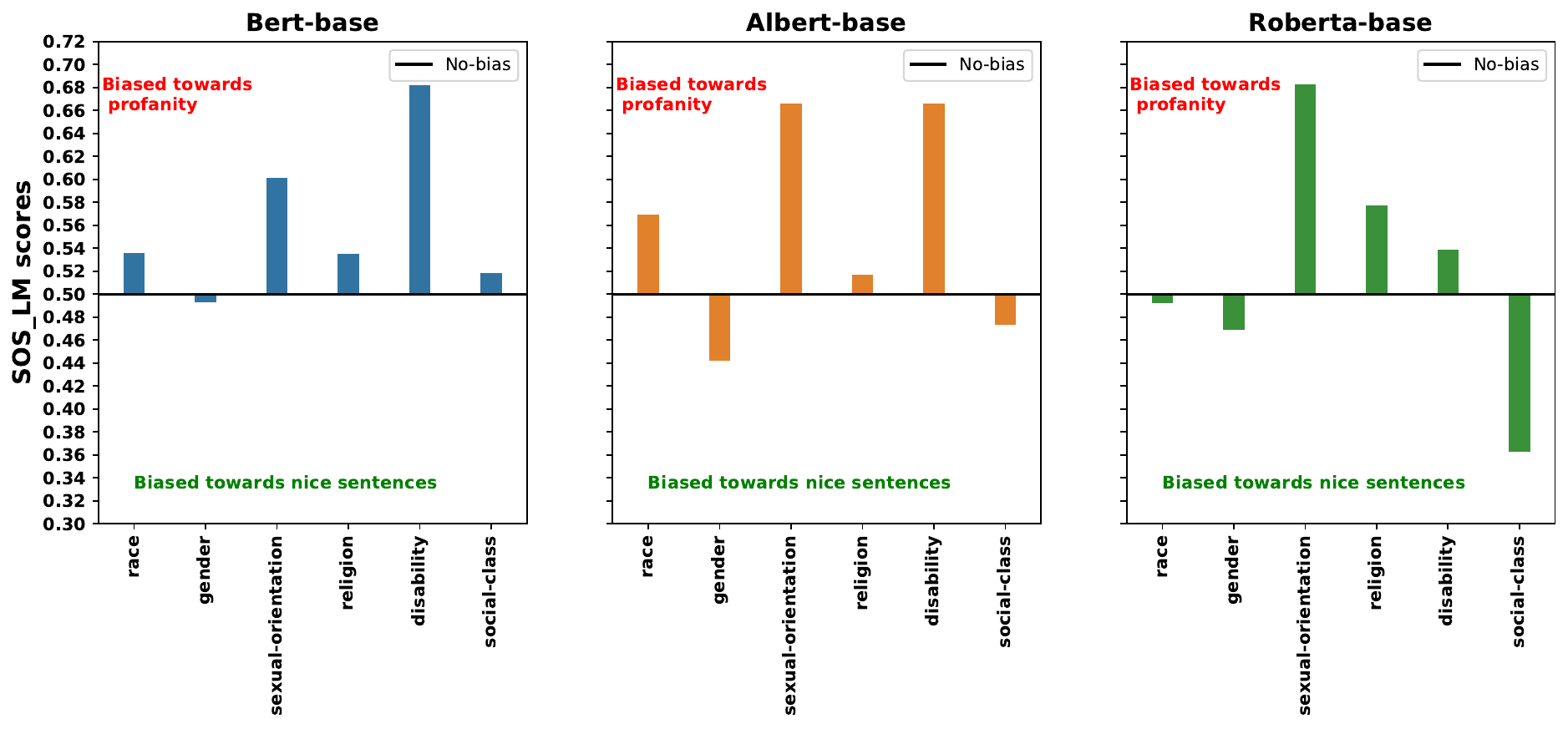}
    \caption{$SOS_{LM}$ bias scores in the different language models \cite{elsafoury2023sosLM}.}
    \label{fig:SOS_LM}
    \end{figure}

I propose a method to measure it and validate it in static \cite{elsafoury_sos_2022} and contextual word embeddings \cite{elsafoury_sos_2022}. Finally, I study how it impacts the performance of these word embeddings on hate speech detection models. I propose the normalized cosine similarity to profanity (NCSP) metric, which is a metric to measure the SOS bias in static word embeddings using the cosine similarity between a list of swear words and non-offensive identity (NOI) words that describe three marginalized groups (Women, LGBTQ, and Non-White) described in \Cref{tab:identity_groups}. As for measuring the SOS bias in contextual word embeddings, I propose the $SOS_{LM}$ metric. The $SOS_{LM}$ metric uses the masked language model (MLM) task to measure the SOS bias, similar to the work proposed in StereoSet \cite{nadeem-etal-2021-stereoset} and CrowS-Pairs \cite{nangia-etal-2020-crows} metrics. Instead of using crowdsourced sentence pairs that express socially biased sentences and socially unbiased sentences, I use synthesized sentence pairs that express profane sentences and non-profane (nice) sentence-pairs.
\begin{table}[t]
\centering
    \renewcommand{\arraystretch}{1.2}
     \resizebox{0.5\textwidth}{!}{
\begin{tabular}{l|l|l}
\hline
Attribute & Marginalized                                                                                                                                   & Non-marginalized                                                                                                                                                                   \\ \hline
Gender              & \begin{tabular}[c]{@{}l@{}}woman, female, girl, wife,\\ sister, daughter, mother\end{tabular}                                                  & \begin{tabular}[c]{@{}l@{}}man, male, boy, son,\\ father, husband, brother\end{tabular}                                                                                            \\ \hline
Race                & \begin{tabular}[c]{@{}l@{}}african, african american,\\ asian, black, hispanic, latin,\\ mexican, indian, \\ middle eastern, arab\end{tabular} & \begin{tabular}[c]{@{}l@{}}white, caucasian, european, \\ american, european, norwegian, \\ german, australian, english, \\ french, american, swedish,  \\canadian, dutch\end{tabular} \\ \hline
Sexual-orientation  & \begin{tabular}[c]{@{}l@{}}lesbian, gay, bisexual,\\ transgender, tran,\\ queer, lgbt,lgbtq,homosexual\end{tabular}                     & hetrosexual, cisgender                                                                                                                                                             \\ \hline
Religion            & \begin{tabular}[c]{@{}l@{}}jewish,buddhist,sikh,\\  taoist, muslim\end{tabular}                                                                & catholic, christian, protestant                                                                                                                                                    \\ \hline
Disability          & blind, deaf, paralyzed                                                                                                                         &                                                                                                                                                                                    \\ \hline
Social-class        & \begin{tabular}[c]{@{}l@{}}secretary, miner, worker, \\ machinist, nurse, hairstylist, \\ barber, janitor, farmer\end{tabular}                 & \begin{tabular}[c]{@{}l@{}}writer, designer, actor, \\ Officer, lawyer, artist,\\ programmer, doctor, \\ scientist, engineer, architect\end{tabular}                              \\ \hline
\end{tabular}}

    \caption{\small{The non-offensive identity (NOI) words used to describe the marginalized and non-marginalized groups in each sensitive attribute. For the disability-sensitive attributes, we use only words to describe disability due to the lack of words used to describe able-bodied.}}
    \label{tab:identity_groups}
\end{table}
I measure the SOS bias scores in 15 static word embeddings \cite{elsafoury_sos_2022} and 3 contextual word embeddings \cite{elsafoury2023sosLM}. The results show that for static word embeddings, there is SOS bias in all the inspected word embeddings, and it is significantly higher towards marginalized groups as shown in \cref{tbl:bias_Scores_marginalised_groups}. Similarly, \Cref{fig:SOS_LM} show that all the inspected contextual word embeddings are SOS biased, but the SOS bias scores are not always higher towards marginalized groups. Then, I validate the SOS bias itself by investigating how reflective it is of the hate that the same marginalized groups experience online. The correlation results, using Pearson correlation coefficient, indicate that there is a positive correlation between the measured SOS bias in static and contextual word embeddings and the published statistics of the percentages of the marginalized groups (Women, LGBTQ, and non-white ethnicities) that experience online hate \cite{hawdon2015online} and the measured SOS bias scores in static word embeddings using the NCSP metric and the $SOS_{LM}$ metric. 
\begin{table}[th]
\renewcommand{\arraystretch}{1.2}
\centering
\resizebox{0.7\columnwidth}{!}{
\begin{tabular}{l|c|c|c}
\hline
\multirow{2}{*}{\textbf{Word embeddings}} & \multicolumn{3}{c}{\textbf{Mean SOS}} \\
\cline{2-4}
 & Women & LGBTQ & Non-white \\ 
\hline
W2V                    & 0.293                                & \textbf{0.475}                       & 0.456                               \\ \hline
Glove-WK               & 0.435                                & \textbf{0.669}                       & 0.234                               \\ \hline
glove-twitter          & \textbf{0.679}                       & 0.454                                & 0.464                               \\ \hline
UD                     & 0.509                                & \textbf{0.582}                       & 0.282                               \\ \hline
Chan                   & \textbf{0.880}                       & 0.616                                & 0.326                               \\ \hline
Glove-CC               & \textbf{0.567}                       & 0.480                                & 0.446                               \\ \hline
Glove-CC-large         & 0.318                                & 0.472                                & \textbf{0.548}                      \\ \hline
FT-CC            & 0.284                                & \textbf{0.503}                       & 0.494                               \\ \hline
FT-CC-sws   & 0.473                                & 0.445                                & \textbf{0.531}                      \\ \hline
FT-WK          & 0.528                                & \textbf{0.555}                       & 0.393                               \\ \hline
FT-WK-sws & \textbf{0.684}                       & 0.656                                & 0.555                               \\ \hline
SSWE                   & 0.619                                & 0.438                                & \textbf{0.688}                      \\ \hline
Debias-W2V             & 0.205                                & 0.446                                & \textbf{0.471}                      \\ \hline
P-DeSIP                & 0.266                                & \textbf{0.615}                       & 0.354                               \\ \hline
U-DeSIP                & 0.266                                & \textbf{0.616}                       & 0.343                               \\ \hline
\end{tabular}}
\caption{\small{The {mean SOS} bias score of each static word embeddings towards each marginalized group. Bold scores reflect the group that the static word embeddings is most biased against \cite{elsafoury_sos_2022}.}}
\label{tbl:bias_Scores_marginalised_groups}
\end{table}
I also validate the proposed metric to measure the SOS bias in comparison to the social bias metrics proposed in the literature. I use the Pearson correlation coefficient between the social bias scores and the SOS bias scores in the static and the contextual word embeddings. The results show that, for the inspected static word embeddings, the correlation results, according to Pearson correlation, show a negative correlation between the measured SOS bias scores measured using the NCSP metric and the social bias scores (gender and race) measured using the WEAT, RND, RNSB, and ECT metrics. As for the contextual word embeddings, the Pearson correlation coefficient results show a positive correlation between the SOS bias scores measured using the $SOS_{LM}$ metric and the social bias scores (gender, race, and religion) measured using the CrowS-Pairs metric, which could be the case because the $SOS_{LM}$ metric is built on the CrowS-Pairs metric.

Finally, I investigate whether the inspected SOS bias explained the performance of the inspected word embeddings on the task of hate speech detection. I train MLP and Bi-LSTM models with an untrainable layer of the different static word embeddings on four hate-speech-related datasets. As for contextual word embeddings, I fine-tune BERT-base-uncased, ALBERT-base, and ROBERTA-base on six hate speech related datasets. Then, I use Pearson's correlation between the SOS bias scores in the different word embeddings and their F1 scores on the models on the task of hate speech detection. The correlation results, similar to the results in \cref{sec:explainabiity}, show an inconsistent positive correlation. This could be because the limitations of other social bias metrics in the literature are extended to the proposed metrics. In this case, the impact of the SOS bias in static and contextual word embeddings on their performance on the task of hate speech detection remains inconclusive.
\paragraph{Contributions:}
The main findings and contributions of the offensive stereotyping perspective can be summarized as follows: \textbf{1)} I define the SOS bias, propose two metrics to measure it in static and contextual word embeddings, and demonstrate that SOS bias correlates positively with the hate that marginalized people experience online. 
\textbf{2)} The results of this section provide evidence that all the examined static and contextual word embeddings are SOS biased. This SOS bias is significantly higher for marginalized groups in  static word embeddings versus non-marginalized groups. However, this is not the case with the contextual word embeddings.
\textbf{3)} Similar to social bias, there is no strong evidence that the SOS bias explains the performance of the different word embeddings on the task of hate speech detection. 
\paragraph{Limitations:}
The findings of this work are limited to the examined word embeddings, models, and datasets, and might not generalize to others. Similarly, the SOS bias scores measured using the NCSP metric in the inspected static word embeddings, are limited to the used word lists. Another limitation is regarding my definition of the SOS bias, as I define bias from a statistical perspective, which lacks the social science perspective as discussed in \citet{Blodgett-etal-2021-norweigan-salmon, DBLP:conf/naacl/DelobelleTCB22}.  Moreover, I only study bias in Western societies where Women, LGBTQ and Non-White ethnicities are among the marginalized groups. However, marginalized groups could include different groups of people in other societies. I also only use datasets and word lists in English, which limits our study to the English-speaking world. Similar to other works on quantifying bias, our proposed metric measures the existence of bias and not its absence \cite{DBLP:conf/naacl/MayWBBR19}, and thus low bias scores do not necessarily mean the absence of bias or discrimination in the word embeddings. Another limitation of this work is the use of template sentence-pairs to measure the SOS bias in contextual word embeddings, which do not provide a real context that might have impacted the measured SOS bias. Since the proposed method used to measure the SOS bias in contextual word embeddings ($SOS_{LM}$) builds on social bias metrics like CrowS-Pairs and StereoSet, it is highly likely that $SOS_{LM}$ have the same limitations as CrowS-Pairs and StereoSet that are pointed out in \citet{Blodgett-etal-2021-norweigan-salmon}. 

\section{The fairness perspective}
\label{sec:fairness}
In \citet{elsafoury2023bias}, I investigate how different sources of bias in NLP models and their removal impact the fairness of the task of hate speech detection. Improving the fairness of the text classification task is very critical to ensure that the decisions made by the models are not based on sensitive attributes like race or gender.

I first measure three sources of bias according to \cite{shah-etal-2020-predictive, hovy2021five}: representation bias, selection bias, and overamplification bias. Then, I fine-tune three language models: BERT, ALBERT, and ROBERTA on the Jigsaw dataset \cite{Jigsaw-tox}, and measure the fairness of these models using two sets of fairness metrics: threshold-based and threshold-agnostic. The threshold-based metrics are the TPR\_gap and the FPR\_gap metrics used in \citet{steed-etal-2022-upstream, De-Arteaga-etal-2019-bias-in-bios}. As for the threshold-agnostic metric, I use the AUC\_gap metric, which is an adaptation of the metrics proposed in \citet{Borkan-etal-2019-naunced-metrics}. I investigate the impact of the different sources of bias on the models' fairness by measuring the Pearson correlation coefficient between the bias scores and the fairness score.  Then, I investigate the impact of removing the three sources of bias, using different debiasing methods, on the fairness of hate speech detection models. I remove the representation bias using the SentDebias method proposed in \citet{liang2020towards} to remove gender, racial, religious and SOS bias on the inspected language models. To remove the selection bias, I aim to balance the ratio of positive examples between the identity groups in the Jigsaw dataset. To achieve that, I generate synthetic positive examples using existing positive examples in the Jigsaw training dataset, but with word substitutions using the NLPAUG tool that uses contextual word embeddings to generate word substitutions \cite{ma2019nlpaug}.  To remove the overamplification bias, I aim to ensure that the different identity groups, in the Jigsaw dataset, appear in similar semantic contexts in the training dataset, as proposed in \citet{webster2020measuring}. To achieve that, I use different methods: 1) create data perturbations, 2) I use the sentDebias method to remove the bias representations from the fine-tuned models.
\begin{table}
\centering
    \renewcommand{\arraystretch}{0.9}
    \resizebox{0.5\textwidth}{!}{
    \begin{tabular}{l|r|r|r}
    \hline
         & \multicolumn{3}{c}{SenseScore}  \\ \hline
        Model & Gender & Race & Religion \\ \hline
        \textbf{ALBERT-base}                                  & $6.9 e^{-05}$ & 0.032 & 0.006 \\ \hline
        + downstream-perturbed-data             & \color{teal}$\downarrow4.2 e^{-05}$ & \color{teal}$\downarrow0.002$ & \color{teal}$\downarrow0.001$ \\ \hline
        + downstream-stratified-data            & \color{red}$\uparrow0.042$ & 0.032 & \color{red}$\uparrow0.009$ \\ \hline
        + downstream- stratified-perturbed-data & \color{red}$\uparrow0.013$ & \color{teal}$\downarrow0.003$ & \color{teal}$\downarrow0.0007$ \\ \hline

        \textbf{BERT-base} & 0.001 & 0.03 & 0.001 \\ \hline
        + downstream-perturbed-data & \color{teal} $\downarrow0.0007$ & \color{teal} $\downarrow0.003$ & 0.001 \\ \hline
        + downstream-stratified-data & \color{red} $\uparrow 0.025$ & \color{teal} $\downarrow0.022$ & \color{red}$\uparrow0.004$ \\ \hline
        + downstream- stratified-perturbed-data & \color{red}$\uparrow0.002$ & \color{teal} $\downarrow0.002$ & \color{teal} $\downarrow0.0008$ \\ \hline

        \textbf{RoBERTa-base} & 0.001 & 0.024 & 0.003 \\ \hline
        + downstream-perturbed-data & \color{teal}$\downarrow0.0008$ & \color{teal} $\downarrow0.006$ & \color{teal} $\downarrow0.001$ \\ \hline
        + downstream-stratified-data &  \color{red} $\uparrow 0.038$ & \color{red} $\uparrow 0.036$ & 0.003 \\ \hline
        + downstream- stratified-perturbed-data &  \color{red} $\uparrow 0.003$ & \color{teal}$\downarrow0.002$ & \color{teal}$\downarrow 0.0003$ \\ \hline
        \end{tabular}}
    \caption{\small{SenseScores of the difference models before and after the different debiasing methods. (\textcolor{red}{$\uparrow$}) means that the extrinsic bias score increased and the fairness worsened.(\textcolor{teal}{$\downarrow$}) means that the extrinsic bias score decreased and the fairness improved \cite{elsafoury2023bias}.}}
    \label{tab:sensescores}
\end{table}
Thereafter, I compare the fairness of the inspected language models on the task of hate speech detection before and after removing each of the inspected source of bias. I aim to find the most impactful source of bias on the fairness of the task of hate speech detection and to find out the most effective debiasing method. The results suggest that overamplification and selection bias are the most impactful on the fairness of the task of hate speech detection and removing it using data perturbations is the most effective debiasing method. I also use the counterfactual fairness method Perturbation score sensitivity ($SenseScore$), proposed in \citet{prabhakaran-etal-2019-perturbation} to further inspect the impact of removing different sources of bias and the most effective bias removal method. The results in \Cref{tab:sensescores} support the results removing overamplification bias is the most effective on improving the fairness of hate speech detection.

Finally, I build on the findings of this work and propose practical guidelines to ensure the fairness of the task of text classification and showcase these recommendations on the task of sentiment analysis.
\paragraph{Contributions:}
The main findings and contributions of the fairness perspective can be summarized as follows: 
\textbf{1)} The results demonstrate that the dataset used to measure the models' fairness on the downstream task of hate speech detection plays an important role in the measured fairness scores. \textbf{2)} The results indicate that it is important to have a fairness dataset with similar semantic contexts and ratios of positive examples between the identity groups within the same sensitive attribute, to make sure that the fairness scores are reliable. \textbf{3)} Unlike the findings of previous research \cite{cao2022, kaneko2022}, the results demonstrate that there is a positive correlation between representation bias, measured by the CrowS-Pairs and the $SOS_{LM}$ metrics, and the fairness scores of the different models on the downstream task of hate speech detection. \textbf{4)} Similar to findings from previous research, \cite{steed-etal-2022-upstream}, the results of this work demonstrate that downstream sources of bias, overamplification and selection, are more impactful than upstream sources of bias, representation bias. \textbf{5)} The results also demonstrate that removing overamplification bias by training language models on a dataset with a balanced contextual representation and similar ratios of positive examples between different identity groups, improved the models' fairness consistently across the sensitive attributes and the different fairness metrics, without sacrificing the performance. \textbf{6)}  I provide empirical guidelines to ensure the fairness of the text classification.
\paragraph{Limitations:}
It is important to point out that the work done in this section is limited to the examined models and datasets. This work studies bias and fairness from a Western perspective regarding language (English) and culture. There are also issues regarding the datasets that those metrics used to measure the bias \cite{Blodgett-etal-2021-norweigan-salmon}. The used fairness metric, extrinsic bias metrics, also received criticism \cite{hedden2021statistical}. This means that even though I used more than one metric and different methods to ensure that our findings are reliable, the results could be different when applied to a different dataset. It is also important to mention that there is a possibility that the findings regarding the most effective debiasing method, which is fine-tuning the models on a perturbed dataset, is the case because I use a perturbed fairness dataset as well. I recognize that the provided recommendations to have a fairer text classification task rely on creating perturbations for the training and the fairness dataset. It might be challenging for some datasets, especially if the mention of the different identities is not explicit, like using the word \say{Asian} to refer to an Asian person but using Asian names instead. Additionally, for the sentiment analysis task, the used keyword to filter the IMDB dataset and get only gendered sentences might provide additional limitations that might have influenced the results. Moreover, in this section, I aim to achieve equity in the fairness of the task of text classification between the different identity groups. However, equity does not necessarily mean equality, as explained in \citet{Broussard2023}.

\section{What have we learned?}
\label{sec:important_lessons}
In this section, I combine all the findings of my thesis and point out how this work can benefit the NLP community and the ongoing research on hate speech detection, bias, and fairness in NLP. The survey of the literature on hate speech detection in \cref{sec:hate_speech_survey} shows a lack of research on the impact of bias in NLP models and hate speech detection models. Especially the impact on the performance of hate speech detection, and how the hateful content led NLP models to form an offensive stereotyping bias, in addition to limitations with the current research that investigates the impact of bias in NLP models on the fairness of hate speech detection models. The aim of my thesis is to fill these research gaps.

The research goal of my thesis is to investigate the bias in NLP models  and its impact on the performance and fairness of the task of hate speech detection, and more generally, the task of text classification. The findings of my thesis show that the bias in NLP models is preventing us from having reliable and effective hate speech detection and text classification models. This is evident by the findings of my thesis.

From the \textbf{Explainability,} perspective, it is inconclusive that the social bias in NLP models explains the performance of hate speech detection models due to limitations in the proposed metrics to measure social bias. However, the results in \cref{sec:explainabiity} also indicate that the bias resulting from pre-training language models, e.g., syntactic bias and biased pre-training datasets, impacts and explains their performance on hate speech detection modes. This good performance suggests that the hate speech detection model associates hateful content with marginalized groups. This might result in falsely flagging content written by marginalized groups on social media platforms.

From the \textbf{Offensive stereotyping bias} perspective,  the findings in \cref{sec:sos} demonstrate that word embeddings, static and contextual, are systematic offensive stereotyping (SOS) biased. The results show no strong evidence that the SOS bias explains the performance of the word embeddings on the task of hate speech detection, due to limitations in the proposed metrics to measure the SOS bias. However, the existence of SOS bias might have an impact on the hate speech detection models in ways that we have not explored or understood yet, especially against the marginalized groups. 

From the \textbf{Fairness} perspective, the findings of \cref{sec:fairness} show that the inspected types of bias, representation, selection, overamplification, have an impact on the fairness of the models on the task of hate speech detection, especially the downstream sources of bias which are selection and overamplification bias. This means that the bias in the current hate speech datasets and the bias in the most commonly used language models have a negative impact on the fairness of hate speech detection models. Hence, researchers should pay attention to these biases and aim to mitigate them before implementing hate speech detection models. 

These findings assert the notion that bias in NLP models negatively impacts hate speech detection models and that, as a community, we need to mitigate those biases so that we can ensure the reliability of hate speech detection models. However, in \cref{sec:bias_survey}, I discuss the limitations and criticisms of the currently used methods to measure and mitigate bias in NLP models that fail to incorporate findings from the social sciences. 

As a short-term solution to improve the fairness of hate speech detection and text classification tasks, I provide a list of guidelines in \citet{elsafoury2023bias}. These guidelines can be summarized as follows:
\begin{enumerate}[wide=0pt]
\itemsep0em 
    \item Measure the bias in the downstream task.
    \item Remove overamplification bias.
    \item To reliably measure fairness, use a balanced fairness dataset and counterfactual fairness metrics.
    \item Choose a model with an acceptable trade-off between performance and fairness.
\end{enumerate}

For a long-term solution and to overcome the current limitations of studying bias and fairness in NLP models, I provide a detailed actionable plan in \citet{elsafoury2023origins} and I summarize the main items in this plan here:

\begin{enumerate}[wide=0pt]
\itemsep0em 
    \item Raise the NLP researchers' awareness of the social and historical context and the social impact of development choices. 
    \item Encourage specialized conferences and workshops on reimagining NLP models with an emphasis on fairness and impact on society.
    \item Encourage specialized interdisciplinary fairness workshops between NLP and social sciences.
    \item Encourage diversity in NLP research teams.
    \item Incorporating more diversity workshops in NLP conferences.
    \item Encourage shared tasks that test the impact of NLP systems on different groups of people.
   \end{enumerate}

\section{Future work}
\label{sec:future_work}
In this section, I discuss important future research directions to mitigate the limitations of this work and the literature on NLP.
\subsection{Widening the study of bias in NLP }
One of the main limitations of this work and most of the work on bias and fairness in NLP models is that it focuses on the English language and on bias from a Western perspective. A critical future work is to create biased datasets in different languages to investigate social bias in models that are pre-trained on data in different languages. It is also important to investigate bias in multilingual NLP models and bias against marginalized groups in societies apart from Western societies.

\subsection{Investigate the impact of social bias causes on the bias in NLP }
In this work, I argue that the sources of bias on the NLP pipelines originate in social sources. I also argue that the methods proposed to measure and mitigate bias in NLP models are inefficient, as a result of failing to incorporate social sciences literature and methods. One of the main limitations of this work is the lack of studies that empirically support this argument. This research direction is an important step towards understanding the bias and fairness in NLP and machine learning models in general.

\subsection{Studying the impact of bias on NLP tasks using causation instead of correlation}
In this work, the measured correlation between sources bias in NLP models and the performance and fairness of NLP downstream tasks, is mostly statistically insignificant. Using causation instead of correlation to investigate that impact could be more effective.
\section{Conclusion}
In this paper, I provide a summary of my PhD thesis. I describe the work done to each my research findings and contributions. I also discuss the limitations of my work and how they can be mitigated in future research. Moreover, I discuss the main lessons learned from my research as well as recommendations that can benefit the NLP research community, especially for studying and mitigating bias in NP models and improving the fairness of text classification tasks.

\bibliography{custom}
\bibliographystyle{acl_natbib}

\end{document}